# Leveraging Interview-Informed LLMs to Model Survey Responses: Comparative Insights from AI-Generated and Human Data

Jihong Zhang[1], Xinya Liang[1], Anqi Deng[2], Nicole Bonge[1], Lin Tan[3], Ling Zhang[4] and Nicole Zarrett[5]

[1]Department of Counseling, Leadership, and Research Methods, University of Arkansas

[2]Department of Health, Human Performance, & Recreation, University of Arkansas

[3]Department of Psychological Science, University of Arkansas

[4]College of Education, University of Wyoming

[5]Department of Psychology, University of South Carolina

## Author Note

Jihong Zhang  https://orcid.org/000-0003-2820-3734

Xinya Liang  https://orcid.org/0000-0002-2453-2162

Anqi Deng  https://orcid.org/0000-0001-7112-3268

Nicole Bonge  https://orcid.org/0009-0003-0609-6576

Lin Tan  https://orcid.org/0000-0002-9384-264X

Ling Zhang  https://orcid.org/0000-0002-9449-6283

Nicole Zarrett  https://orcid.org/0000-0001-9035-132X

Correspondence concerning this article should be addressed to Jihong Zhang, Department of Counseling, Leadership, and Research Methods, University of Arkansas, 109 Graduate Education Building, Fayetteville, AR 72703, Email: jzhang@uark.edu

**Funding Support:** Research reported in this publication was supported by the National Institute of Nursing Research of the National Institutes of Health under Award Number 1R01NR017619-01 (Zarrett, PI). The content is solely the responsibility of the authors and does not necessarily represent the official views of the National Institutes of Health.

Acknowledgements: We would like to thank our school and after school program community partners.

**Disclosure Statement**: The authors report there are no competing interests to declare.

**Abstract**

Mixed methods research integrates quantitative and qualitative data but faces challenges in aligning their distinct structures, particularly in examining measurement characteristics and individual response patterns. Advances in large language models (LLMs) offer promising solutions by generating synthetic survey responses informed by qualitative data. This study investigates whether LLMs, guided by personal interviews, can reliably predict human survey responses, using the Behavioral Regulations in Exercise Questionnaire (BREQ) and interviews from after-school program staff as a case study. Results indicate that LLMs capture overall response patterns but exhibit lower variability than humans. Incorporating interview data improves response diversity for some models (e.g., Claude, GPT), while well-crafted prompts and low-temperature settings enhance alignment between LLM and human responses. Demographic information had less impact than interview content on alignment accuracy. Item-level analysis revealed higher discrepancies for negatively worded questions, suggesting LLMs struggle with emotional nuance. Person-level differences indicated varying model performance across respondents, highlighting the role of interview relevance over length. Despite replicating individual item trends, LLMs faltered in reconstructing the test's psychometric structure. These findings underscore the potential of interview-informed LLMs to bridge qualitative and quantitative methodologies while revealing limitations in response variability, emotional interpretation, and psychometric fidelity. Future research should refine prompt design, explore bias mitigation, and optimize model settings to enhance the validity of LLM-generated survey data in social science research.

**Leveraging Interview-Informed LLMs to Model Survey Responses: Comparative Insights from AI-Generated and Human Data**

**Introduction**

Mixed methods design is a widely used research approach in psychology and education (Bishop, 2015; Johnson et al., 2007; Powell et al., 2008) due to the complementary strengths of integrating both quantitative and qualitative approaches. The framework of mixed methods study design requires the rigorous collection and analysis of both quantitative and qualitative data to answer research questions and test hypotheses (Creswell & Clark, 2017). However, because of the structural differences between qualitative (e.g., open-ended responses, textual richness, context-dependent interpretations) and quantitative data (e.g., numeric ratings, standardized scales, fixed-response options; Schoonenboom, 2023), comparative analysis between quantitative and qualitative information—especially for examining measurement characteristics such as response formatting or question wording, as well as person-level attributes such as individual response styles and subjective interpretations—remains challenging, which hinders the broader adoption of mixed methods design (J. Wang et al., 2024). With the advent of large language models (LLMs) in natural language understanding (NLU) tasks, bridging quantitative and qualitative insights has become increasingly feasible.

Generative Artificial Intelligence (GenAI), particularly LLMs, enables researchers to analyze survey data, automate scoring in educational assessment, and conduct large-scale social simulations by generating synthetic personas (A. Li et al., 2025). Personas are generalized representations of targeted users based on real users' data. The underlying assumption is that, when provided with relevant context (e.g., a scoring rubric, demographic information), LLMs can simulate personas given personal information (e.g., role information of teachers, survey respondents, interview coders, or demographic information) to represent population user profiles, which can produce population-level opinions that approximate those of real-world target populations. Researchers can then use these synthetic opinions to predict real-world responses (e.g., human scoring, public opinions on elections). For instance, prior studies have relied on census data to generate personas that reflect varying political ideologies, which were subsequently used to predict survey responses on political attitudes and improve measurement accuracy by analyzing discrepancies between predicted and observed data (Argyle et al., 2023; Chang et al., 2024; Yu et al., 2024).

Prior research in this area has primarily focused on improving the representation of LLM-generated responses for various populations to improve the generalizability of the LLMs' output. This process often relies on large sample sizes in the training process of LLMs (A. Li et al., 2025). However, in real-world settings, the vast majority of studies in education and psychology routinely operate with relatively small sample sizes (Slavin & Smith, 2009), a challenge that remains underexplored in LLM literature. In addition, although Likert-scale surveys are widely used in the social sciences, limited research has examined the performance of LLMs in generating valid Likert-scale responses (Liu, Sharma, et al., 2024) and to our knowledge, no research has examined the alignment between LLM-generated responses and human survey responses when rich individual-level interview informs the LLMs.

In light of the growing interest in incorporating qualitative insights with LLM-driven survey methods, the present study examines whether LLMs can extract relevant information from personal interview data and reliably predict individuals' survey responses. We illustrate this approach using the *Behavioral Regulations in Exercise Questionnaire* (BREQ; Cid et al., 2012) and interviews from after-school program staff as an example. This study aims to explore whether LLMs can leverage the human interview to generate scale responses that closely align with those provided by humans. By comparing survey responses generated by interview-informed LLMs to observed human survey responses, we can then identify inconsistencies that may shed light on important information about measurement or human respondents. Through the innovative application of LLMs to interpret qualitative interview data and generate corresponding quantitative psychometric responses, this research introduces a novel approach to bridge between qualitative and quantitative methodologies.

This study is organized as follows: the *Background* section reviews prior work on LLMs to generate survey responses; the *Current Study* section outlines the research purpose and guiding questions; the *Study Design* section describes the data, experimental settings, and LLM implementation; the *Results* section compares LLM-generated outputs with human survey data; and the *Discussion* section explores the empirical implications and suggests directions for future research.

**Background**

LLMs have been investigated in educational and psychological assessment for varied purposes (May et al., 2025). Recent studies have demonstrated the versatility of LLMs in extracting insights from unstructured texts, either by treating unstructured texts as sequences of natural language processing (NLP) tasks (Parker et al., 2024) or leveraging multi-agent frameworks for natural language understanding (NLU) and simulation (Rasheed et al., 2024). However, P. Wang et al. (2024) caution that although LLMs effectively capture broad patterns, they often struggle with item-level nuances and subtle individual differences. Additionally, the methods and effectiveness of using LLMs to generate human-like quantitative responses remain inadequately examined.

*Persona-Driven Method*

The most widely used approach to simulating survey responses with LLMs centers on the persona-driven method (Jiang et al., 2023; A. Li et al., 2025; Liu, Sharma, et al., 2024). *Synthetic persona generation* (Chen et al., 2024) involves creating artificial user profiles for applications such as political attitude modeling and economic forecasting. These synthetic personas are typically generated by sampling demographic attributes (e.g., age, race, income) from census or survey distributions. Other data-driven techniques of persona generation include clustering respondent data, conducting factor analysis, or applying matrix decomposition methods to identify archetypal personas from extensive population datasets (Jansen et al., 2022). An alternative approach, termed *descriptive personas*, does not rely on real-world demographic attributes. Instead, descriptive personas utilize synthetic text inputs to condition LLMs to directly generate diverse personas. These inputs are commonly derived from extensive corpora of public web texts, allowing the models to simulate a broader spectrum of social perspectives and behavioral traits (Ge et al., 2024). This study utilizes the descriptive persona generation method by combining research context, real-world personal interview data, and demographic information.

*Application of LLM-Generated Responses*

In addition to studies on persona generation, researchers have devoted increasing attention to using LLMs to generate human-like quantitative survey responses for various purposes (Liu, Sharma, et al., 2024; Xu & Zhang, 2023). These LLM-generated responses offer a cost-effective solution for data augmentation or prediction, reducing the need for time-intensive human data collection. A key area of debate concerns the LLMs' assumed "personality" in simulating responses. Some studies posit that LLMs can emulate underlying psychological traits, enabling researchers to mimic real-world human respondents with LLM-generated samples (Huang, Wang, et al., 2024; Y. Li et al., 2024a, 2024b; Liu et al., 2025; Serapio-García et al., 2023). In this context, LLMs' generated responses are viewed as *silicon samples* (also referred to as "synthetic datasets") processing emergent personality characteristics shaped by prompts and training data (Sarstedt et al., 2024; Sun et al., 2024). From this standpoint, LLMs may serve as substitutes for human participants under certain controlled conditions.

Conversely, other studies minimize the personality traits of LLMs, instead conceptualizing LLMs as efficient computational tools for measurement-development tasks, such as item calibration (Ding et al., 2024), automated scoring (Mendonça et al., 2025; Uto & Uchida, 2020), one-on-one tutoring (Fateen & Mine, 2025), cheating detection (Q. Wang & Li, 2025), and item generation (Laverghetta Jr. & Licato, 2023). This perspective positions LLMs more closely to traditional computational software systems than to simulated agents with underlying psychological constructs. This contrast, viewing LLMs as synthetic respondents or task-specific tools, reflects LLMs' versatile applications in AI research.

*LLM Variants and Configurations*

Key factors that may influence the performance of LLM-generated survey responses include the choice of LLM-based chatbot, temperature settings, and prompt configurations. LLM-based chatbots are AI chatbots developed and trained by different companies. The current literature provides limited comparative analyses of these chatbots in the context of generating educational or psychological survey responses. Many existing studies employ only a single chatbot for simulation purposes. However, prior research suggests that investigations into LLM capabilities should extend beyond a single model (e.g., GPT) to include other AI chatbots for robust comparative analysis (e.g., Agarwal et al., 2023; Lozić & Štular, 2023; S. Wu et al., 2023). The release of GPT-4 by OpenAI in 2023 attracted global attention due to its strong test-taking performance (Nori et al., 2023; OpenAI et al., 2023). Similarly, Claude 2, released by Anthropic in June 2023, gained interest for its extended context window (up to 100,000 tokens), significantly expanding the working memory of LLMs. Gemini, developed by Google, has also demonstrated high performance on medical benchmarks (Saab et al., 2024). Furthermore, several open-source LLMs have proven effective across domains. In the context of survey response generation (commonly referred to as silicon sampling), widely used chatbots include GPT models (e.g., GPT-3 Turbo, GPT-4, Sarstedt et al., 2024) and Llama models (e.g., Llama-3, Peng et al., 2024).

LLM parameters and prompt design represent two additional important factors in LLM simulation studies. For instance, Sarstedt et al. (2024) discuss how model parameters (e.g., temperature) and prompt tuning affect the quality of silicon samples. Temperature is a parameter of LLMs used to control the degree of randomness when choosing tokens. A temperature of 0 yields highly deterministic outputs by consistently selecting the highest-probability tokens, resulting in minimal variation across runs. In contrast, higher temperatures introduce greater variability and creativity in generated responses. However, it is to note that an excessively high temperature could degrade response quality faster than it adds originality. In addition, prompt-tuning is an important process of LLM-based generation that may impact outputs. Minor changes in the prompt can split the difference between the AI tool failing to understand the instruction and outperforming the request (Ekin, 2023; Federiakin et al., 2024). Neither temperature nor prompt settings are well examined in the survey response generation literature.

**Current Study**

The current study's purpose is to examine the quality of LLM-based chatbot-generated quantitative survey responses based on qualitative interview data within the context of social sciences. This study has three aims: (1) to understand the variability of LLM-generated survey responses across LLM chatbots, temperature settings, and prompt settings; (2) to identify key factors that affect alignment between LLM-generated and human responses; (3) to explore measurement-level and person-level characteristics by comparing LLM-generated responses to human responses. In this study, measurement-level and person-level characteristics (e.g., demographics, interview length, item wordings etc.) are defined as item-level and person-level information relevant to measurement and respondents that may have impacts on the alignment between LLM-generated responses and human responses.

To achieve these goals, we designed a simulation study to assess the feasibility of using LLMs as an evaluation tool. This study utilizes the real-world interview data and survey responses collected from adult afterschool program (ASP) personnel who were participating in the Connect through PLAY program, a randomized controlled efficacy trial (Author's Masked), designed to support changes in the physical activity (PA) values, motivations, and behavior of ASP staff/teachers as a means for establishing sustainable social changes in the school setting for increasing the daily PA of underserved youth. The mixed methods research design that involved collecting complimentary survey and interview data made it an ideal data set to test this study's primary aims. The research questions of this study are as follows:

1. How do LLMs' generated survey responses compare with human responses in terms of means and variability of item responses across different settings, such as LLM chatbots, temperature settings, and prompt configurations?

2. How do factors such as LLM chatbots, prompt settings, and temperature settings influence the alignment between LLM-generated and human responses?

3. What do discrepancies between LLM-generated and human responses indicate about measurement and person characteristics?

## Method

**Data**

This study employed the Behavioral Regulation in Exercise Questionnaire (BREQ; Mullan et al., 1997; Mullan & Markland, 1997; Wilson et al., 2002) and semi-structured interviews as the primary instruments to assess health-based behavioral regulation among after-school program (ASP) staff and program directors (N = 55) across 10 ASPs serving underserved youth in the Southeastern United States from 2023 to 2024. Specifically, qualitative data were collected through semi-structured interviews conducted between researchers and ASP staff, each lasting approximately 15–25 minutes, and subsequently transcribed for analysis. The staff semi-structured interview instrument is designed to assess ASP staff's physical activity (PA) experiences, perceptions, and readiness to implement new programming, particularly in underserved communities (see interview questions in the Supplementary). It captures key constructs including community engagement, job satisfaction, motivation towards PA, and perceived support; staff understanding of youth experiences and approaches to youth development; beliefs and attitudes about their physical and mental health; and perceptions of stress among both youth and staff. Additionally, it explores readiness for future program implementation by identifying potential motivators, barriers, and support needs, including professional development, incentives, and access to resources. In addition to the interviews, ASP staff were asked to complete a series of self-report questionnaires, including the BREQ(Mullan et al., 1997; Mullan & Markland, 1997; Wilson et al., 2002). The BREQ instrument is used to assess the motivation behind exercise behavior. For the present analysis, data included responses from 19 ASP staff members who completed both the interview and the questionnaire ($N = 19$). All interview transcripts were de-identified by removing or masking personal information prior to analysis.

**Measures**

*The Behavioral Regulations in Exercise Questionnaire (BREQ)*

BREQ (Mullan et al., 1997; Mullan & Markland, 1997; Wilson et al., 2002) is a 15-item survey measuring the self-determination in exercises. Each item of BREQ was rated on a 6-point Likert scale: 1 = Strongly disagree, 2 = Disagree, 3 = Somewhat disagree, 4 = Somewhat agree, 5 = Agree, 6 = Strongly agree. The BREQ contains four subscales: a four-item subscale of external regulation (e.g., *I exercise because other people say I should.*), a three-item subscale of introjected regulation subscale (e.g., *I feel guilty when I don't exercise.*), a four-item subscale of identified regulation (e.g., *I value the benefits of exercise.*), and a four-item subscale of intrinsic regulation (e.g., *I exercise because it's fun.*). The BREQ has been widely used in exercise motivation research, and studies have shown the instrument to be a reliable and valid measure of exercise motivation in various populations (Cronbach's $\alpha$ is .81 to .89; Cid et al., 2012).

**Study Design**

In general, the procedure of this study consists of three steps (see Figure 1): (1) data collection; (2) LLM simulation; and (3) comparison and evaluation. In Step 1, all participants were invited to complete a semi-structured interview and structured questionnaires. In Step 2, the collected information and research background information were used to create different prompts. These prompts differed in the components they included, which instructed LLM on how to generate survey responses. Finally, in Step 3, we compared the generated responses and actual survey responses using various evaluation metrics. Description of each prompt and evaluation metric are detailed below.

**Figure 1**

*Procedure of the simulation study*

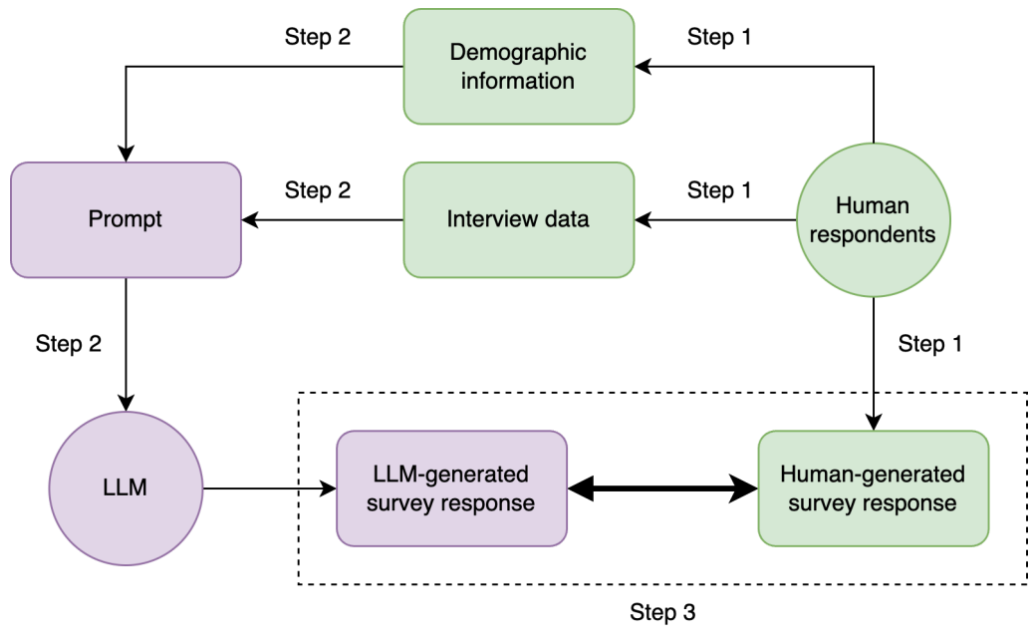

For the LLM simulation, we considered three key design factors: (1) LLM chatbots, (2) temperature setting (Low = 0; High = 0.5), and (3) prompt components. First, we evaluated three widely adopted commercial generative AI models (LLM chatbots): OpenAI's GPT-4.1 (*gpt*), Google's Gemini 2.0 Flash (*gemini*), and Anthropic's Claude 3.7 Sonnet (*claude*). These models were selected given their prominence and accessibility in current applications of large language modeling. Second, we manipulated the temperature parameter to control the degree of randomness in token selection during the response generation. We selected 0.5 as the high-temperature condition based on preliminary analyses indicating that temperatures exceeding 0.5 (e.g., 0.7) produced highly variable and conversational outputs that lacked the structure necessary for generating coherent item-level survey responses (Jiang et al., 2023). Third, as shown in Figure 2, four prompts were developed based on varying combinations of prompt components (see Supplementary materials). These prompt types were designed to evaluate whether the inclusion of specific information components influences the performance of LLM-based response simulation. Specifically, Prompt 1 (*P-BR*; Baseline Research Background & Survey) served as the baseline condition, containing only the research background and survey information. Prompt 2 (*P-BR-PI*; P-BR + Personal Interview) and Prompt 3 (*P-BR-DI*; P-BR + Demographic Information) extended P-BR by incorporating personal interview data or demographic information, respectively. Prompt 4 (*P-BR-PI-DI;* P-BR + Personal Interview + Demographic Information) was the most comprehensive, combining all elements from Prompt 2 (research background, survey item content, interview responses) and demographic information, thus including the richest information and representing the most personalized prompt.

**Figure 2**

*Components of four prompts*

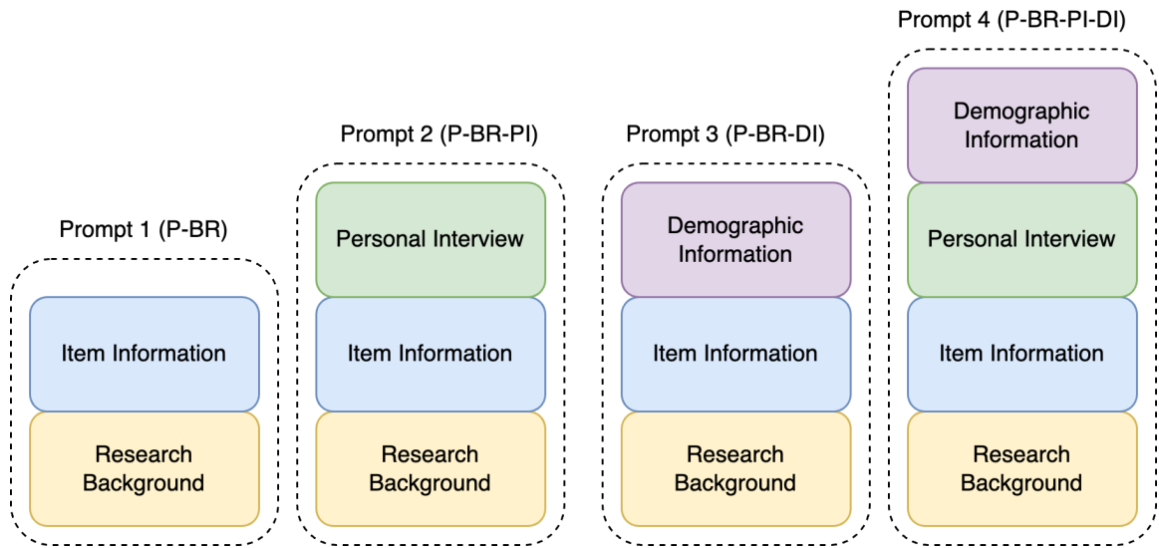

To further evaluate the prompts used in current study, we quantified the amount of information of one prompt by the number of tokens used for each prompt following the *o200k_base* encoding used by GPT-4o. The results showed that Prompt 1 has the lowest number of tokens (all prompts have same number of tokens as 1,276), Prompt 2 and Prompt 3 have middle level of number of tokens (Mean=5,917; Min = 3,190; Max=9,641), and Prompt 4 has the highest number of tokens across all samples (Mean=5,952; Min=3,227; Max=9,676), suggesting Prompt 4 has the highest amount of information followed by Prompt 2 and Prompt 3, and Prompt 1 has the least amount of information. For each prompt, the LLMs were instructed to simulate the role of the interviewee and generate the response to each item of the BREQ mentioned above.

In total, there are 3 (LLM chatbots) × 4 (prompts) × 2 (temperature settings) = 24 conditions. The calling functions of *Application Programming Interface* (API) of LLMs were created in Python and the data analysis was performed in R. The code is available on the OSF platform (DOI:10.17605/OSF.IO/AFQG3).

**Evaluation criterion**

*Alignment among LLMs*

The alignment among three LLMs (*gpt*, *claude*, and *gemini*) was evaluated using the following methods. First, we calculated and visualized the average item means and variances of LLM-generated responses over participants in each condition. Second, we calculated the average *Pearson* correlations ($\rho$) between LLM-generated responses for each condition. Third, we performed a three-way ANOVA with Pearson's correlations as the dependent variable (DV) and three factors (LLM pairs, prompt settings, and temperature settings) as independent variables (IVs) to identify important factors that influenced the alignment among LLMs.

*Alignment between LLMs and Human*

We first presented descriptive statistics of LLM-generated responses and human-generated responses including item-level means and variances in different conditions to compare LLMs to human respondents. Second, we compared LLM-generated responses to human-generated survey responses in different layers. We utilized item-level (Equation 1), person-level (Equation 2), and test-level root-mean-square deviation (RMSE; Equation 3) to quantify the deviation between LLM-generated and human-generated responses in different aspects. Item- and person-level RMSEs can help identify which items or persons exhibit the largest deviation between LLM-generated responses and human survey responses. These may indicate some person/item characteristics that LLMs struggle to capture or reveal potential measurement errors. The three types of RMSEs are defined as follows.

Item-level RMSE is calculated as the square root of the average squared difference between LLM-generated and human-generated responses for each item across all participants.

$$RMSE_i = \sqrt{\frac{\sum_1^P (X_{i,p,AI} - X_{i,p,Human})^2}{P}} \quad (1)$$

where $X_{i,p,AI}$ is the LLM-generated response for item $i$ from participant $p$, and $X_{i,p,Human}$ is the human-generated response for item $i$ from participant $p$. $P$ denotes the total number of samples.

Person-level RMSE is used to quantify the difference between LLM-generated and human-generated responses for each participant across all items.

$$RMSE_p = \sqrt{\frac{\sum_1^I (X_{i,p,AI} - X_{i,p,Human})^2}{I}} \quad (2)$$

Test-level RMSE is used to evaluate the deviations between LLM-generated total test score and human-generated total test score across all LLM samples and human participants.

$$RMSE_T = \sqrt{\frac{\sum_1^P (RAI_{p,AI} - RAI_{p,Human})^2}{P}} \quad (3)$$

$$RAI_{p,\cdot} = -2 * S_{p,ext} - S_{p,int} + 2 * S_{p,ide} + S_{p,int} \quad (4)$$

where $RAI_{p,\cdot}$ of the BREQ denotes the *relative autonomy index* of person $p$, which can be calculated as the weighted sum of four subscale scores using the formula in Equation 4. $S_{p,ext}$, $S_{p,int}$, $S_{p,ide}$, and $S_{p,int}$ denote the mean scores of external regulation, introjected regulation, identified regulation, and intrinsic regulation, respectively.

## Results

**Descriptive statistics**

As shown in Figure 3, all three LLMs produced approximately similar tendencies of item means compared to human participants across items, temperature settings, and prompts. Specifically, items 1 to 7 had relatively lower item means than items 8 to 15. The similarity of average item means across all samples between Prompt 1 and other prompts may suggest that LLMs may capture the items' linguistical features and potential human' responses to items regardless of the amount of personal information in the prompts. However, across all conditions, LLMs tended to generate more extreme responses than humans; that is, for items with lower human ratings (e.g., items 1–7), LLM mean scores were even lower, and for items with higher human ratings (e.g., items 8–15), LLM mean scores were even higher. In addition, temperature settings did not show much difference in item means across all three LLMs. Prompts only had slight differences in the means for certain LLMs, such as *gemini*, but not for others.

**Figure 3**

*Mean item responses for LLMs and Human*

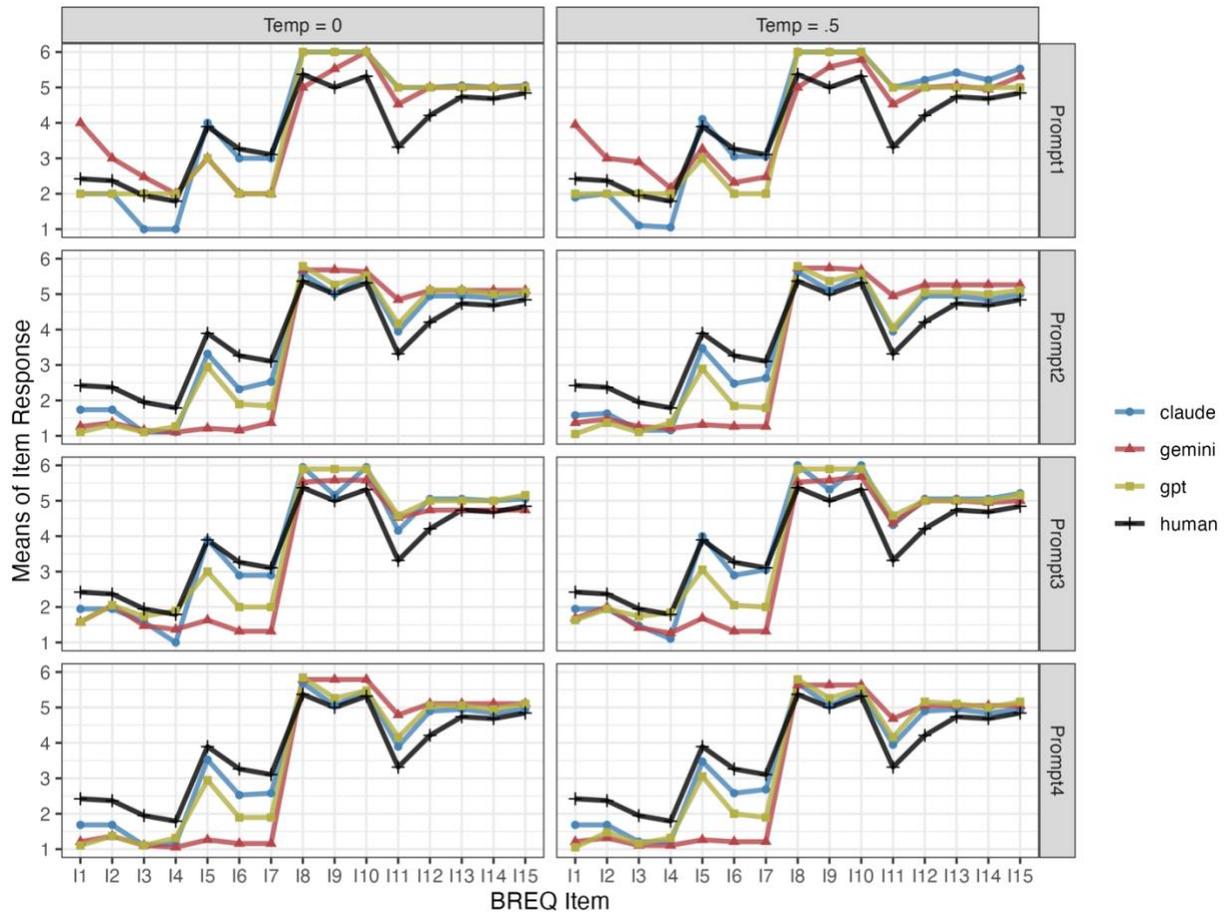

Figure 4 presents the variances of item responses across all respondents or LLM-generated samples in different conditions. Unlike item means, LLM-generated responses showed much lower variability than human responses under all conditions, indicating that LLM-generated responses were less diverse than human responses. That is, the more information one prompt contains, the higher variances LLM-generated responses yielded. Combined with evidence from Figure 3, this indicated that LLMs produced more consistently extreme values. In addition, the variances of LLM-generated responses interacted with LLM chatbots— *claude* had more diverse responses when prompts contained the interview data (Prompt 2 and Prompt 4). In contrast, *gemini* had more diverse responses when prompts contained demographic information only (Prompt 3). In addition, higher temperature conditions (Temp = .5) showed higher variances of item responses than lower temperature conditions (Temp = 0) for all LLM chatbots, which was consistent with the definition of temperature.

**Figure 4**

*Item variances for LLMs and Human*

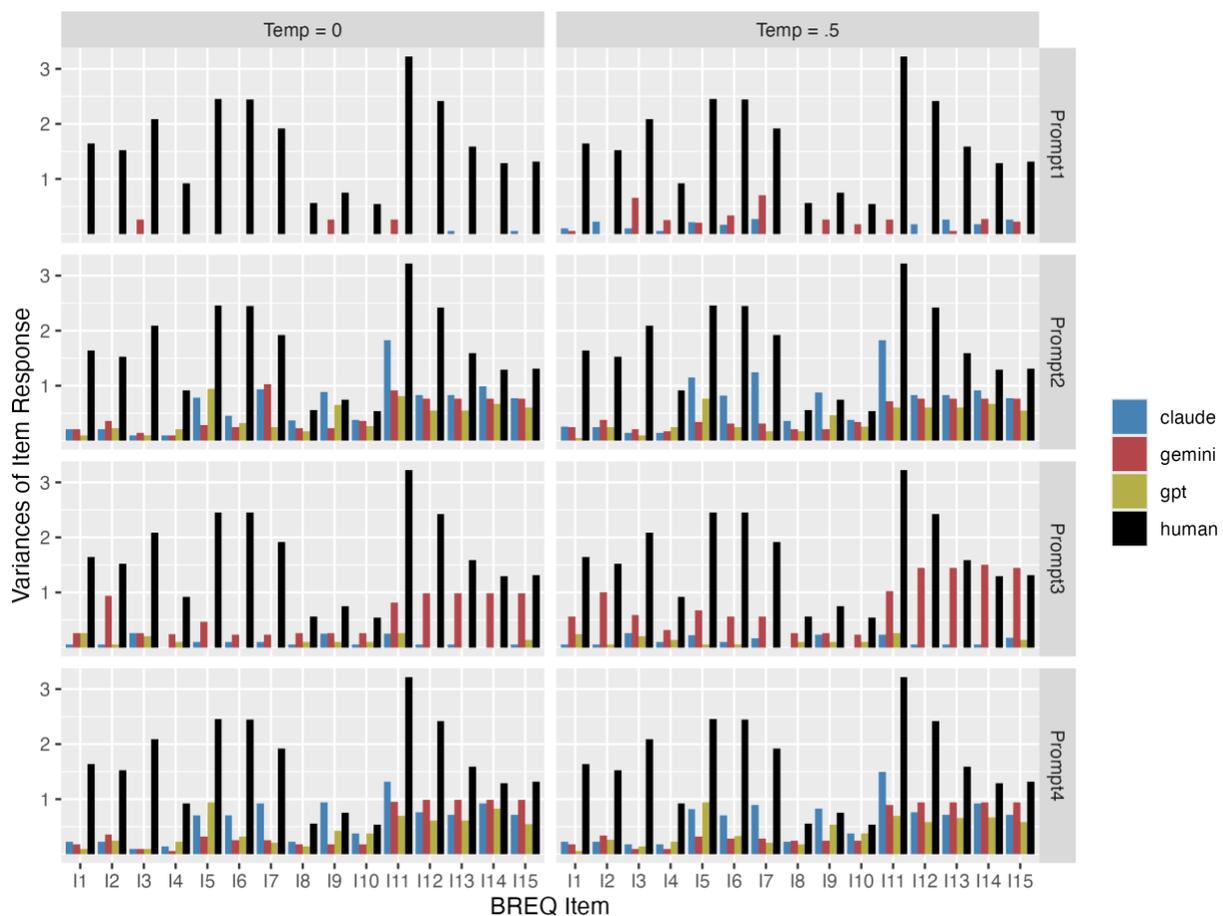

For comparisons among LLMs, the results show that compared to Prompt 1, both *claude* and *gpt* can have higher variances when the prompt contains interview data (Prompt 2 and Prompt 4) while *gemini* can have higher variances when the prompt includes demographic information (Prompt 3 and Prompt 4).

**Alignment among LLMs**

Table 1 presents the Pearson correlations among the three LLMs in different conditions. In each condition, highest correlations were styled with bolded fonts. Results show that *gpt* and *claude* have the highest correlations across various temperatures and prompts ($\rho \in [0.92, 0.95]$), followed by *gpt* and *gemini* ($\rho \in [.87, .93]$), and *claude* and *gemini* ($\rho \in [.81, .88]$). The results also show that the correlations between the three LLM chatbots in the conditions of low temperature (Temp = 0; $\bar{\rho} = .91$) are slightly higher than those in the conditions of higher temperature (Temp = .5; $\bar{\rho} = .89$). In addition, the results show that the correlations in the conditions of prompts containing personal interview data (Prompt 2 and Prompt 3; $\bar{\rho} = .92$) are higher than prompts without personal interview data (Prompt 1 and Prompt 3; $\bar{\rho} = .88$). The results of three-way ANOVAs show that prompt settings ($F_{3,17} = 16.657, p < .001$) and temperature settings ($F_{1,17} = 8.133, p = .011$) have significant main effects on the correlations among three LLM chatbots.

**Table 1**

*Correlations among three LLM chatbots across all conditions*

| Temp | Prompt | $\rho_{gpt,claude}$ | $\rho_{gpt,gemini}$ | $\rho_{gemini,claude}$ |
|---|---|---|---|---|
| Low | Prompt 1 | 0.943 | 0.913 | 0.839 |
| Low | Prompt 2 | 0.953 | 0.925 | 0.882 |
| Low | Prompt 3 | 0.928 | 0.911 | 0.827 |
| Low | Prompt 4 | 0.951 | 0.930 | 0.875 |
| High | Prompt 1 | 0.925 | 0.874 | 0.806 |
| High | Prompt 2 | 0.940 | 0.932 | 0.862 |
| High | Prompt 3 | 0.923 | 0.879 | 0.807 |
| High | Prompt 4 | 0.944 | 0.935 | 0.875 |

**Alignment between LLMs and Humans**

As shown in Figure 5, the average Pearson correlations between LLMs chatbot responses with human responses show that there are medium to high relationships ($\rho \in [.5, .73]$). Among the three LLM chatbots, *claude* shows the highest correlations with humans for all four prompts, followed by *gpt*. In contrast, *gemini* has the lowest correlations with humans across conditions. The results also show that prompts containing interview data (Prompt 2 and Prompt 4) have relatively higher associations with humans than other prompts (Prompt 1 and Prompt 3). Prompt 2 and Prompt 4 have comparable levels of correlations across all LLM chatbots. Prompt 3 (prompt containing demographic information) has slightly higher correlations than the baseline (Prompt 1) for *claude*, *gemini* and *gpt*.

**Figure 5**

*Correlations between LLMs and human responses for four prompts*

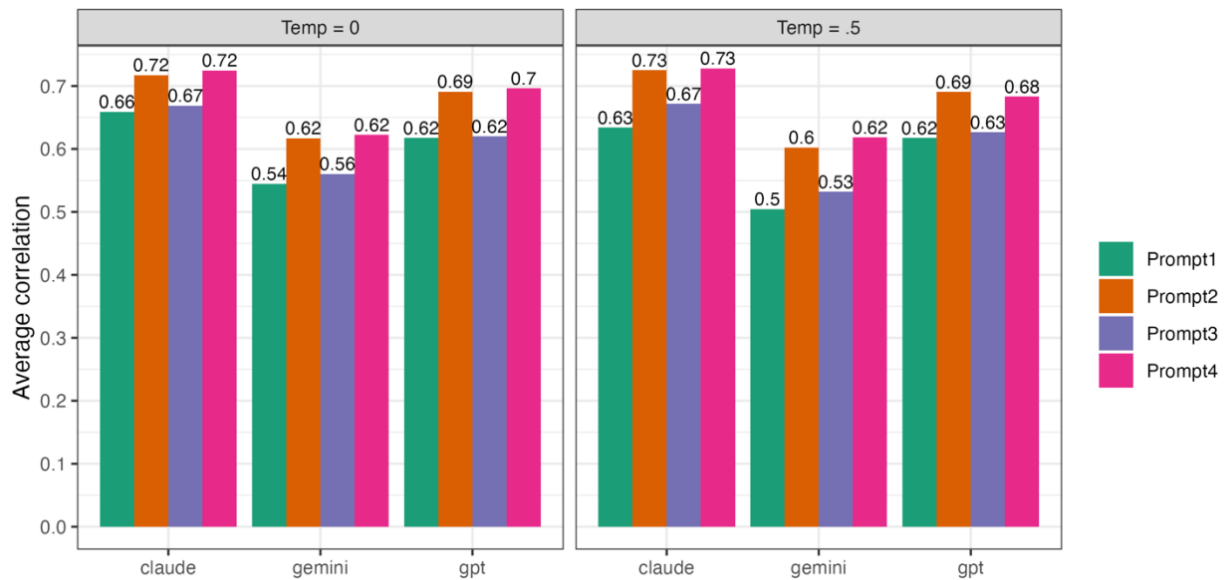

*Item-level RMSEs*

Figure 6 shows the item-level RMSEs for 15 items in different conditions. Overall, *claude* shows lower RMSEs than *gemini* and *gpt*, suggesting that *claude* has the highest alignment with human respondents. In comparison, *gemini* has the highest item-level RMSEs among the three LLM chatbots. As for specific items, we found that items 6, 7, and 11 displayed relatively higher RMSEs than other items. After screening those items carefully, we found that item 6 (*I feel ashamed when I miss an exercise session.*), item 7 (*I feel like a failure when I haven't exercised in a while.*), and item 11 (*I get restless if I don't exercise regularly.*) contain relatively negative emotional words, such as "*ashamed*", "*failure*", and "*restless*" while other items have relatively positive words, such as item 8 (*I value the benefits of exercise.*) or item 10 (*I think it is important to make the effort to exercise regularly.*).

**Figure 6**

*Item-level RMSEs between LLMs and human respondents*

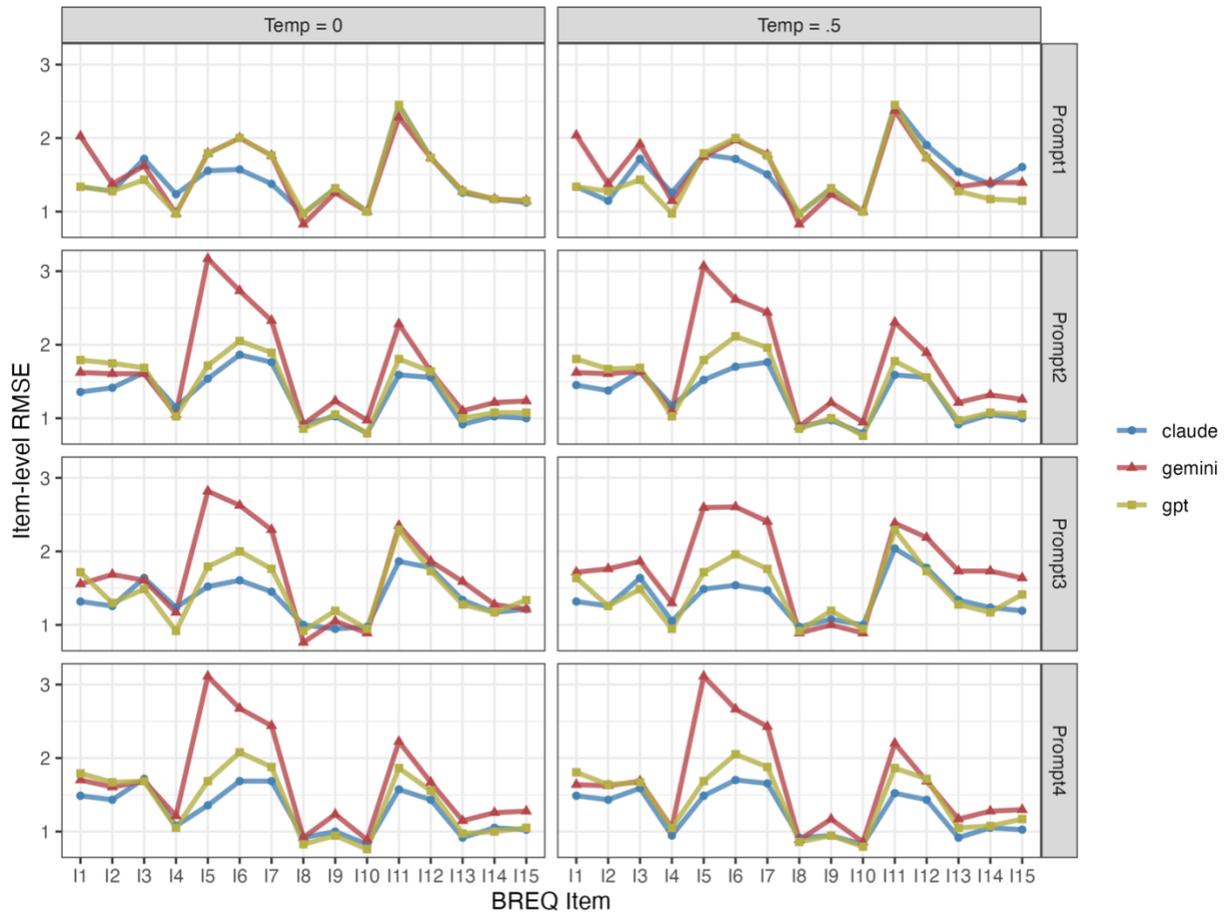

*Person-level RMSEs*

Figure 7 shows the person-level RMSEs between 19 human respondents and LLM-generated samples. When using Prompt 2 and Prompt 4, two respondents (ID 2111 and 2303) have the highest alignment between their survey responses with the LLMs' generated responses, indicated by the lowest person-level RMSEs. *claude* has the lowest person-level RMSEs when using Prompt 2 and Prompt 4, suggesting that it has the highest alignment with human respondents. In addition, the correlation between the number of tokens in individual interviews and the average person-level RMSE for Prompt 2 and Prompt 4 was moderate while not statistically significant ($\rho = 0.404$, $p = .086$), suggesting that longer interviews (greater number of tokens in prompts) did not necessarily lead to better alignment between LLM-generated and human responses. Nonetheless, such an outcome may stem from our small sample size and limited power.

**Figure 7**

*Person-level RMSEs between LLMs and human respondents*

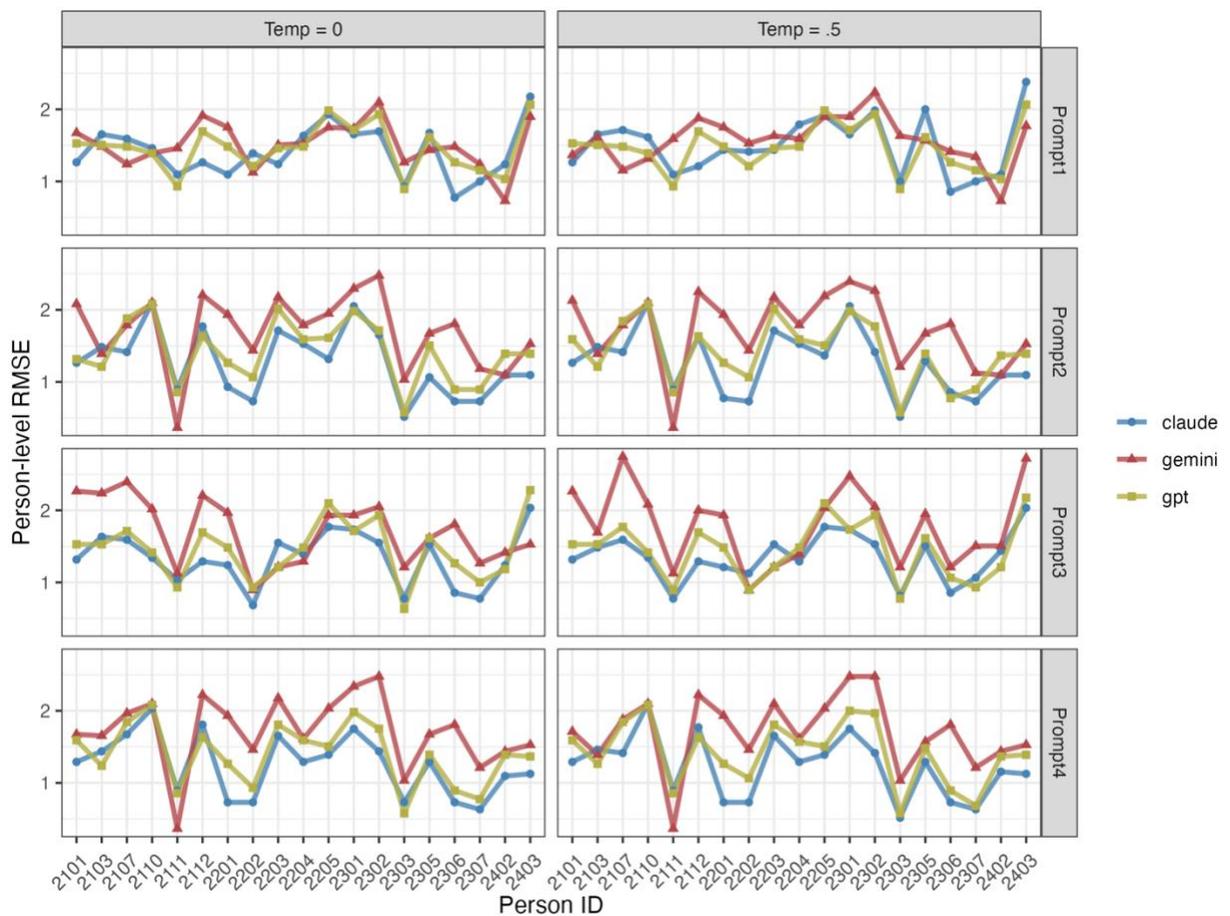

*Test-level RMSEs*

Finally, Figure 8 shows the test-level RMSEs of the *relative autonomy index* (RAI; see Equation 4) for all conditions. The results suggested that compared to Prompt 1, only *claude* with Prompt 2, Prompt 3, and Prompt 4 showed higher alignment with human respondents at the test level, which indicated, for some chatbots (*gemini* and *gpt*), more personal information in the prompts (e.g., interview data and demographic information) may not necessarily improve the alignment between LLMs and humans towards the RAI. The calculation of RAI requires theoretical understanding of the definitions and relationships among study variables (specifically for this study, external regulation, introjected regulation, identified regulation, and intrinsic regulation), which is not included in the training of LLM chatbots.

**Figure 8**

*Test-level RMSEs between LLMs and human respondents*

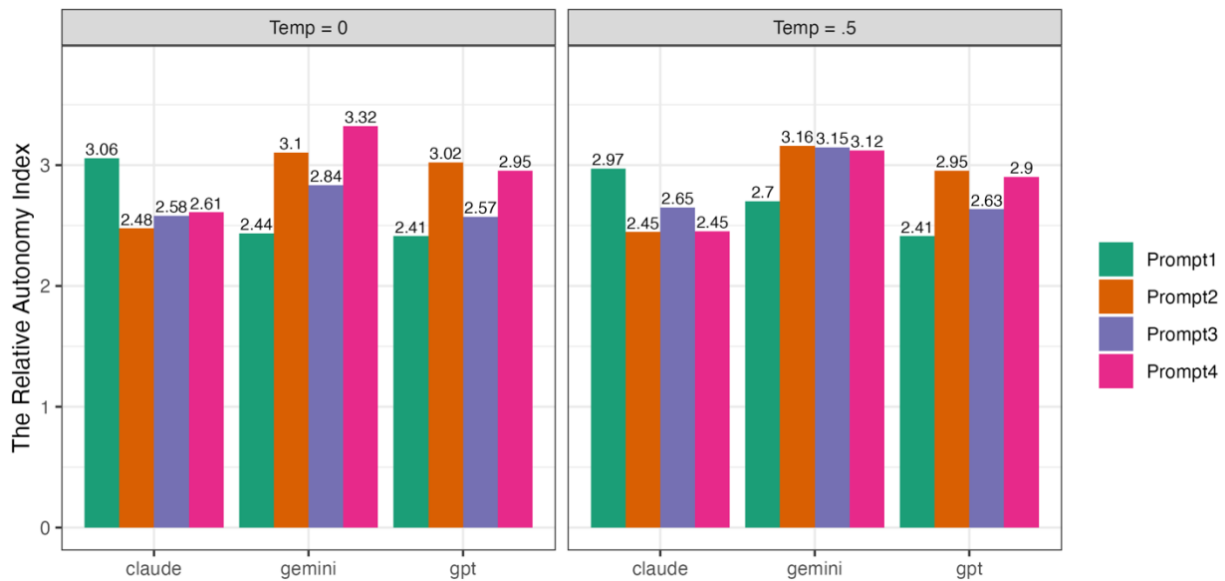

**Discussion**

This study introduces an LLM-driven approach that leverages personal interviews of after-school program staff to simulate quantitative survey responses. The comparisons between LLM chatbots' generated responses and the staff's actual survey data provide indications of variability in accuracy of LLMs' survey responses by measurement and personal characteristics that may help researchers improve their research design.

Our first finding is that post-trained LLM chatbots can capture the overall pattern of BREQ responses from after-school program staff across different temperature and prompt settings; however, LLM-generated responses tend to exhibit lower variability than those of human respondents (see Figure 3 and Figure 4). In general, incorporating personal interview data can improve the diversity of generated responses for some LLM chatbots (i.e., *claude* and *gpt*). This finding is consistent with other LLM-driven simulation studies that well-crafted personas can yield results that approximate real-world human behaviors on average but hardly approximate the variability of human behaviors (e.g., A. Li et al., 2025).

Our second finding is that prompts with interview data (Prompt 2 and Prompt 4) and/or low-temperature settings can improve the alignment among LLMs chatbots. This finding is consistent with prior studies that have shown high-relevant and high-quality prompts are important for the stability of LLM outputs. For example, previous studies found that LLM prompted with high-quality analytic rubrics can yield results that approximate human grading logics in the automated scoring task (e.g., X. Wu et al., 2025).

Our third finding is that prompts incorporating personal interview and other personal information (e.g., demographics) can affect the alignment between LLM-generated responses and human survey responses (see Figure 5). Findings indicated personal interview data can improve the alignment between LLM-generated responses and human responses, but additional demographic information in the prompts may not as notably improve the alignment (Prompt 4 vs. Prompt 2). Surprisingly, individuals' demographic information was less impactful on alignment compared to interview data perhaps because the relationship between individuals' demographic information and their survey responses is relatively weaker than the interview. Further research is needed to investigate the importance of different types of qualitative data in training LLMs for research in social sciences.

Our fourth and final finding is that variations in the alignment between LLMs and human responses indicate further investigation into item and person characteristics is needed. Specifically, the item-level RMSEs between LLM-generated and human responses are higher for survey items with negative emotional words (e.g., *ashamed*, *failure*, and *restless*; see Figure 6), suggesting that LLMs may have difficulty understanding the relationship between negative emotional wording in the interview data and real personal attitudes showed in the survey. However, stronger alignment for items with positive wording does not guarantee optimal item properties. Both LLM-generated and human responses may be extreme (e.g., choosing 1 = Strongly disagree or 6 = Strongly agree), indicating a need to review item wording to prevent such polarized answers. The person-level RMSEs suggest that LLMs differ in their ability to learn from diverse respondents. We found a moderate but not significant association between the length of individual interviews (measured by the number of tokens) and the alignment. This may suggest that the relevance of interview content, rather than its length, plays a more critical role in enhancing the alignment of LLM-generated responses. However, given our limited sample size, the impact of the interview length on the alignment warrants further investigation. The test-level RMSEs suggest that LLMs struggle to grasp the overall psychometric structure of the test when item scores are aggregated and weighted by subscales. In other words, while they can mimic individual item patterns, LLMs have difficulty reproducing the underlying psychometric relationships of the whole test. This finding is consistent with prior studies that LLMs can capture individual differences in personality traits and emotional expressions (Y. Li et al., 2024a; Serapio-García et al., 2023) but may not adhere to psychometric principles rigorously (Huang, Jiao, et al., 2024; Huang, Wang, et al., 2024; A. Li et al., 2025; Liu, Bhandari, et al., 2024; P. Wang et al., 2024).

**Limitations and Future Directions**

*Designing Effective Prompts Using Qualitative Data*

Although prior studies have demonstrated that when guided by appropriately designed prompts, LLMs can serve as a viable method for validating the consistency between qualitative (e.g., interview) and quantitative (e.g., survey) data, the scientific principles underlying prompt construction remain underdeveloped. A central challenge lies in identifying which attributes of respondents should be included in prompts to optimize LLM-generated responses for specific research purposes. In the current study, demographic and interview-based information were incorporated. However, future research could benefit from considering additional respondent characteristics such as psychographic (e.g., personality traits, values), behavioral (e.g., past actions), or contextual variables (e.g., social or interview setting), as suggested by A. Li et al. (2025). Systematic investigations are needed to determine how to tailor these elements for domain-specific applications.

*Limited Diversity in LLM-Generated Responses*

Consistent with prior research, our findings suggest that LLM-generated responses often exhibit lower diversity than actual human respondents. The underlying causes of this limitation, however, remain poorly understood. Further investigation is required to identify whether this constraint is rooted in model architecture, training data, or the input prompt features.

*Bias in Prediction Accuracy Across Individuals*

The accuracy of LLM-generated responses in approximating real individual responses appears to vary across participants and items, raising questions about LLM's bias from measurement and personal characteristics. Previous studies have shown that LLMs are likely to present various types of biases depending on the demographic information of individuals (Binz & Schulz, 2023; Dillion et al., 2023; Yan et al., 2024). One possibility is that prompt content containing strong emotional valence—especially positive sentiment or highly emotional expressions—may lead to distorted outputs, as noted by A. Li et al. (2025). Additionally, emotional inconsistencies across data collection modes (e.g., differing emotional states during interviews vs. surveys) may contribute to mismatches between LLM's predictions and actual responses. If emotional tone or opinion varies across these modalities, LLM's predictions may reflect that divergence. This suggests that interview protocols could be improved by encouraging emotional neutrality or consistent perspectives, potentially enhancing the alignment between qualitative data and LLM-generated survey responses.

*Leveraging Interview Data to Reduce Bias*

More research is needed to explore how researchers can leverage interview data to mitigate bias in LLM-generated outputs. Qualitative data analysis techniques, such as word cloud visualizations or topic modeling, may help identify language features—such as topic richness or information density—that influence the fidelity and variability of generated responses. Understanding these dynamics may offer new pathways for refining prompt design and improving model performance.

*Model Temperature and Response Stability*

Other factors that significantly affect LLM-generated survey responses include the temperature settings of the LLM. Our findings suggest that increasing the temperature may enhance the diversity of generated responses. However, temperatures exceeding a certain threshold (e.g., >0.8) tend to produce outputs that lack coherence and structural integrity. Currently, there is a lack of empirical guidance on optimal temperature settings for survey simulation tasks, particularly in education and psychology. Future research should focus on identifying appropriate model settings that balance diversity with consistency, particularly in high-stake applications such as educational assessment.

## Declaration of Generative AI Software Tools in the Writing Process

During the preparation of this work, the author(s) used ChatGPT in the section Abstract in order to improve clarity and refine language. After using this tool, the author(s) reviewed and edited the content as needed and take(s) full responsibility for the content of the publication.

LLM SURVEY RESPONSES 37Powell, H., Mihalas, S., Onwuegbuzie, A. J., Suldo, S., & Daley, C. E. (2008). Mixed methods research in school psychology: A mixed methods investigation of trends in the literature. *Psychology in the Schools*, *45*(4), 291–309. https://doi.org/10.1002/pits.20296

Rasheed, Z., Waseem, M., Ahmad, A., Kemell, K.-K., Xiaofeng, W., Duc, A. N., & Abrahamsson, P. (2024). Can large language models serve as data analysts? A multi-agent assisted approach for qualitative data analysis. *arXiv Preprint*. https://doi.org/10.48550/arXiv.2402.01386

Saab, K., Tu, T., Weng, W.-H., Tanno, R., Stutz, D., Wulczyn, E., Zhang, F., Strother, T., Park, C., Vedadi, E., & al., et. (2024). Capabilities of gemini models in medicine. *arXiv*. https://doi.org/10.48550/arXiv.2404.18416

Sarstedt, M., Adler, S. J., Rau, L., & Schmitt, B. (2024). Using large language models to generate silicon samples in consumer and marketing research: Challenges, opportunities, and guidelines. *Psychology & Marketing*, *41*(6), 1254–1270. https://doi.org/10.1002/mar.21982

Schoonenboom, J. (2023). The fundamental difference between qualitative and quantitative data in mixed methods research. *Forum Qualitative Sozialforschung / Forum: Qualitative Social Research*, *24*(11). https://doi.org/10.17169/fqs-24.1.3986

Serapio-García, G., Safdari, M., Crepy, C., Sun, L., Fitz, S., Romero, P., Abdulhai, M., Faust, A., & Matarić, M. (2023). Personality traits in large language models. *arXiv Preprint*. https://doi.org/10.48550/arXiv.2307.00184

Slavin, R., & Smith, D. (2009). The relationship between sample sizes and effect sizes in systematic reviews in education. *Educational Evaluation and Policy Analysis*, *31*(4), 500–506. https://doi.org/10.3102/0162373709352369